\documentclass[letterpaper]{article} 
\usepackage{aaai24}  
\usepackage{times}  
\usepackage{helvet}  
\usepackage{courier}  
\usepackage[hyphens]{url}  
\usepackage{graphicx} 
\usepackage{natbib}  
\usepackage{caption} 
\usepackage{algorithm}
\usepackage{algorithmic}

\usepackage{newfloat}
\usepackage{listings}
\DeclareCaptionStyle{ruled}{labelfont=normalfont,labelsep=colon,strut=off} 
\floatstyle{ruled}
\newfloat{listing}{tb}{lst}{}
\floatname{listing}{Listing}
\title{How Real Is Real?\\ A Human Evaluation Framework for Unrestricted Adversarial Examples}
\author{
    Dren Fazlija\textsuperscript{\rm 1},
    Arkadij Orlov\textsuperscript{\rm 2},
    Johanna Schrader\textsuperscript{\rm 1},\\
    Monty-Maximilian Zühlke\textsuperscript{\rm 1},
    Michael Rohs\textsuperscript{\rm 2},
    Daniel Kudenko\textsuperscript{\rm 1}
}
\affiliations{
    \textsuperscript{\rm 1}L3S Research Center, Appelstr. 4, 30167 Hannover, Germany\\
    \textsuperscript{\rm 2}Leibniz University Hannover, Welfengarten 1, 30167 Hannover, Germany\\
    \{dren.fazlija, schrader, zuehlke, kudenko\}@l3s.de,\\ arkadij.orlov@stud.uni-hannover.de, michael.rohs@hci.uni-hannover.de
}

\usepackage{bibentry}

\newcommand{\approach}{\textsc{Scooter}} 
\usepackage{soul}
\usepackage{xcolor}

\usepackage{amsmath}

\begin{document}

\maketitle

\begin{abstract}
With an ever-increasing reliance on machine learning (ML) models in the real world, \textit{adversarial examples} threaten the safety of AI-based  systems such as autonomous vehicles. 
In the image domain, they represent maliciously perturbed data points that look benign to humans (i.e., the image modification is not noticeable) but greatly mislead state-of-the-art ML models. 
Previously, researchers ensured the imperceptibility of their altered data points by \textit{restricting} perturbations via $\ell_p$ norms. 
However, recent publications claim that creating natural-looking adversarial examples without such restrictions is also possible. 
With much more freedom to instill malicious information into data, these \textit{unrestricted} adversarial examples can potentially overcome traditional defense strategies as they are not constrained by the limitations or patterns these defenses typically recognize and mitigate. This allows attackers to operate outside of expected threat models. 
However, surveying existing image-based methods, we noticed a need for more human evaluations of the proposed image modifications. 
Based on existing human-assessment frameworks for image generation quality, we propose \approach{} -- an evaluation framework for unrestricted image-based attacks.
It provides researchers with guidelines for conducting statistically significant human experiments, standardized questions, and a ready-to-use implementation. 
We propose a framework that allows researchers to analyze how imperceptible their unrestricted attacks truly are.
\end{abstract}

\section{Introduction}

Since 2013, it has been well-established that malicious entities can mislead sophisticated computer vision models by adding imperceptible noise to images~\cite{szegedy2013intriguing}. 
The resulting \textit{adversarial examples} (AEs) look benign to humans.
However, due to the restricted nature of the attacks, their effectiveness can be limited significantly through image pre-processing~\cite{dziugaite2016study} and certified robust defense strategies~\cite{li2020sok}. 
Hence, \textit{unrestricted} AEs have garnered increasingly more interest in the last few years. 
Such attacks alter images significantly through modifications that humans easily overlook,
which is achieved by taking semantic information into account.
For example,~\cite{shamsabadi2020colorfool} perform significant color changes only to non-sensitive areas (e.g., furniture) that look natural to the human eye across a wide range of colors.
Works on unrestricted attacks should rigorously assess the imperceptibility of the resulting images, as this characteristic can no longer be assumed. 
However, few publications employ human evaluation experiments to support their claim – none of which offer statistically significant insights. 
Hence, providing the research community with a statistically significant human evaluation protocol based on well-established study design recommendations is crucial. 



\textbf{Contributions.}
To facilitate research into unrestricted adversarial examples, we propose \approach{} (\textbf{S}ystemizing \textbf{C}onfusion \textbf{O}ver \textbf{O}bservations \textbf{T}o \textbf{E}valuate \textbf{R}ealness) -- a human evaluation framework for examining the quality of unrestricted adversarial images (i.e., the \textit{imperceptibility} of modifications). 
Drawing inspiration from existing tools~\cite{otani2023toward} and following study design recommendations~\cite{aguinis2021mturk}, \approach{} enables researchers to make statistically significant claims about the imperceptibility of image-based attacks. 
The \approach{} framework encompasses $(i)$ a ready-to-use web application with a modular design allowing researchers to integrate their images easily;
$(ii)$ a carefully crafted study protocol that guides researchers in every step of performing online studies;
$(iii)$ an online leaderboard enabling the comparisons of state-of-the-art attacks across different target models; 
$(iv)$ an image database containing all generated AEs for further analyses.

\section{State-Of-The-Art Assessment Protocols}

The most similar work to ours is the evaluation protocol of~\cite{otani2023toward} for analyzing the quality of Text-To-Image generators. 
The authors provide researchers with well-designed domain-specific questions and user interfaces, recommendations for several design choices (e.g., requirements that participants need to fulfill), and templates for reporting human evaluation results. 
While our goals align with the authors', their protocol does not sufficiently guide inexperienced researchers in difficult aspects of experiment design. 
Most notable is the lack of methods to guarantee high-quality evaluation data. 
For example, their protocol does not cover standard measures like attention and instruction manipulation checks. 
The researchers also publicly state their eligibility requirements, which is not recommended as it increases the self-misrepresentation of participants ~\cite{aguinis2021mturk, bauer2020review}. 
In contrast to their work, we aim to support inexperienced researchers by rigorously defining every detail of the study design. 
Another adjacent publication~\cite{zhou2019hype} provides a basic framework for collecting human image quality assessments.
While the resulting protocols HYPE$_{\text{time}}$ and HYPE$_{\infty}$ are widely used in subjective image quality assessment tasks, they display similar weaknesses to ~\cite{otani2023toward}.
We build on the insights of both publications while considering these flaws.

\section{Framework Design}

\begin{figure}[h!]
    \centering
    \includegraphics[width=0.9\columnwidth]{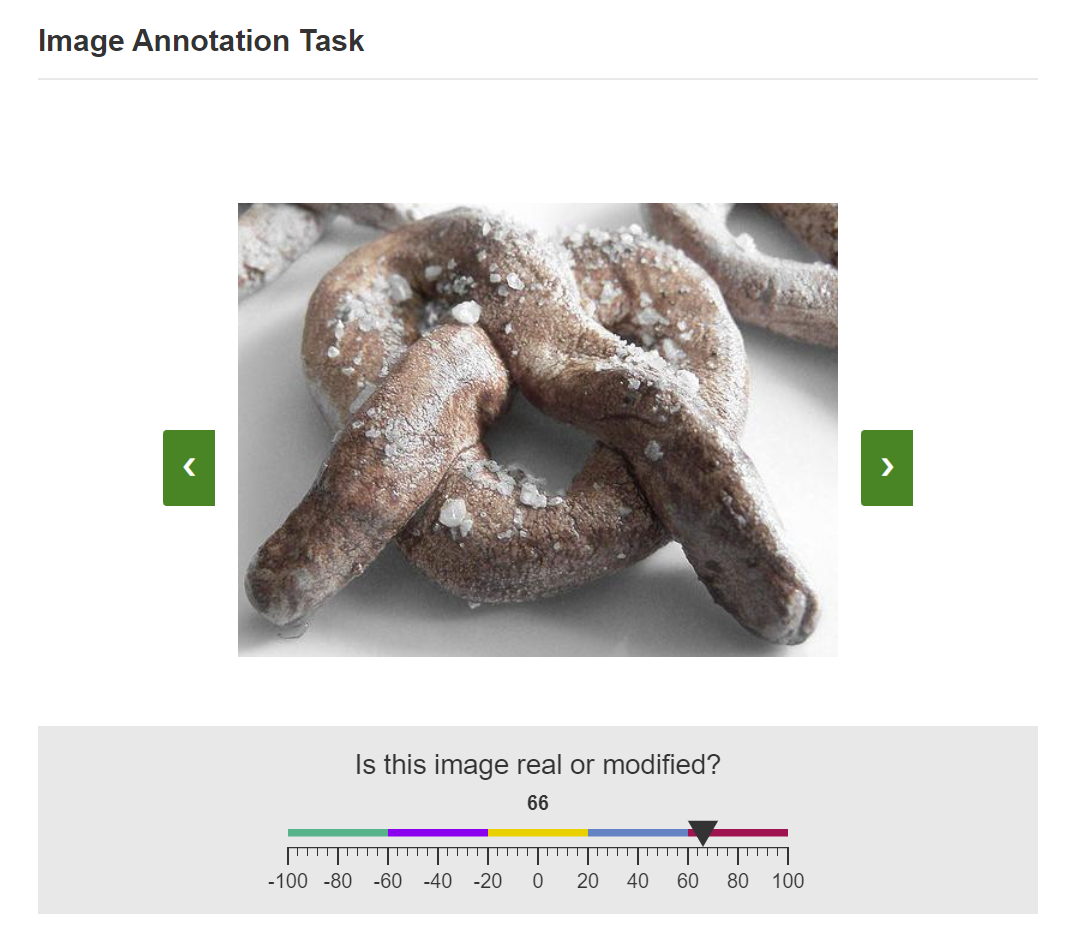}
    \caption{Prototype of the Main Study Interface}
    \label{fig:sliderinput}
\end{figure}

\textbf{Online Study Design.}
To evaluate the imperceptibility of unrestricted attacks, we propose to conduct a 13-minute online study on Prolific\footnote{https://www.prolific.com/}.
We use Prolific because it can prescreen participants without publicly sharing eligibility requirements while providing higher-quality data than other services~\cite{douglas2023data}.
We use the built-in prescreeners to filter out workers who are colorblind and those who self-report to be not fluent in English. 
We further ensure workers' capabilities by performing a short colorblindness and comprehension check. 
In line with Prolific compensation guidelines and related publications, we offer an average compensation of £7.60 per hour. 
To support our protocol's critical design decisions (e.g., the slider-based input as in Figure~\ref{fig:sliderinput}), we developed a Flask-based\footnote{https://flask.palletsprojects.com/} web application to perform these studies. 
Research groups can use the web app to replicate our study design for their experiments. 

\begin{figure}[h!]
    \centering
    \includegraphics[width=0.9\columnwidth]{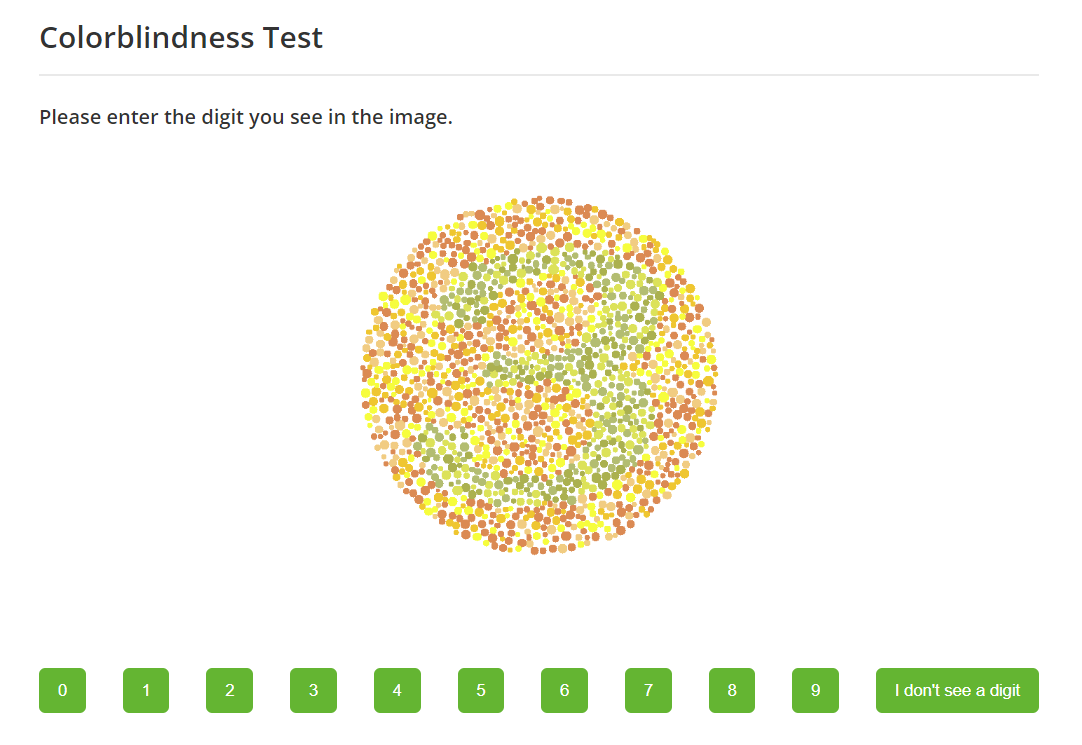}
    \caption{Prototype of the Colorblindness Check Interface}
    \label{fig:ishiharaexample}
\end{figure}

\textbf{Colorblindness Check.}
The most prominent attack vectors for unrestricted AEs are the colors of an image. 
As such, colorblind annotators will likely overestimate the imperceptibility of most attacks. 
Thus, participants must correctly classify five different Ishihara-like images~\cite{ishihara1918tests} before accessing the study's central portion (see Figure~\ref{fig:ishiharaexample}).
These images emulate so-called Ishihara plates, which are widely used diagnostic tools for vision deficiencies. 
Four images show a digit, while one image always displays no digit. 
Failing this check will end the study, and the user will be compensated for 1 minute of work.

\begin{figure}[h!]
    \centering
    \includegraphics[width=0.9\columnwidth]{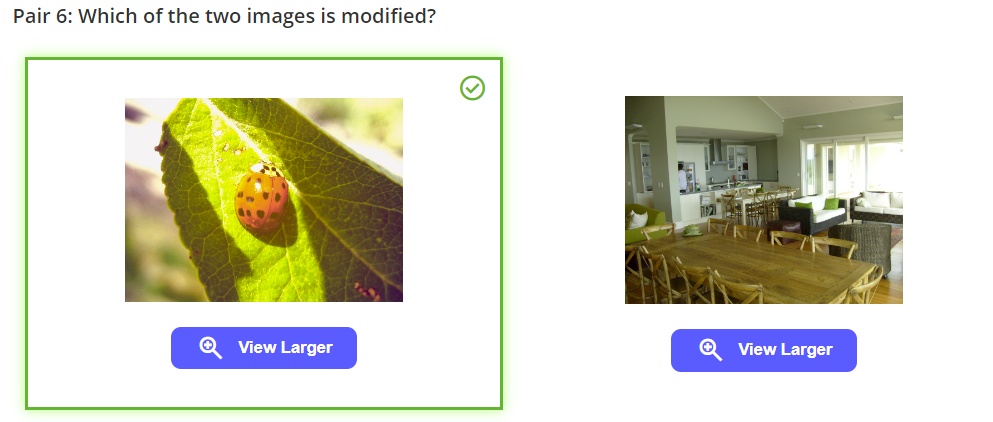}
    \caption{Prototype of the Comprehension Check Interface}
    \label{fig:ccexample}
\end{figure}

\textbf{Comprehension Check.}
After passing the colorblindness check, we provide users with a brief explanation outlining common modification strategies. 
Examples include image filters and the change of colors in a particular area of the image.
After reading the explanation, the user must pass a comprehension check. 
We display six image pairs, each containing a random unmodified and one random modified image (see Figure~\ref{fig:ccexample}). 
To move on to the main portion of the study, participants must correctly classify the modified image of at least five pairs. 
Failing this check will end the study, and the user will be paid for 2.5 minutes of work.

\textbf{Main Study.}
Here, we aim to analyze the imperceptibility of an attack strategy for one specific victim model. 
We evaluate generated AEs by asking users to rate the degree of modification for 106 images, 50 of which are unmodified ImageNet~\cite{imagenet15russakovsky} samples, that the victim classifies correctly with high confidence. 
Another 50 images represent random AEs generated by the assessed attack. 
The remaining six images represent attention checks to ensure the attentiveness of participants. 
Instead of a forced binary choice between "modified" and "unmodified", we want users to rate how confident they feel about the degree of modification via a slider input (see Figure~\ref{fig:sliderinput}).

The input ranges from -100, \textit{I am 100\% certain that this image is unmodified}, to +100, \textit{I am 100\% certain that this image is modified}.
Collecting continuous ratings from many participants will lead to better-informed comparisons between unrestricted attacks, as such ratings can capture finer nuances of attack perceptibility~\cite{chyung2018evidence}. 
For instance, attack A, which produces successful attacks with imperceptible modifications, could be rated the same as attack B, whose successful modifications "barely convince" the average user.

\textbf{Empirical Sample Size Estimation.}
A key factor for conducting statistically significant studies is the choice of an appropriate sample size. 
However, we cannot perform an apriori sample size estimation due to this domain's lack of an established effect size. 
Hence, we must perform studies to empirically determine a suitable number of participants per (attack, model) pair. 
For this purpose, we plan to collect large amounts of data for three attacks on one single victim model. 
Given 3,000 modified and 3,000 unmodified images per (attack, model) pair, which we distribute across 60 unique datasets, we want to collect at least ten samples per image. 
Hence, we must invite at least 600 people per (attack, model) pair. 
To create a recommended buffer of 15\% for low-quality annotations~\cite{aguinis2021mturk}, we will invite another 90 annotators per attack-model pair. 
Using the adversarially trained ResNet-50~\cite{hed2015resnet} model of~\cite{salman2020adversarially}, we will collect data from 690 annotators for three attacks, which vary in complexity.
Based on these results, we will then be able to determine the sufficient (and ideally much smaller) sample size needed for analyzing unrestricted attacks.
\section{Conclusions}
Unrestricted image-based attacks will play a significant role in the near future, especially considering the rapid progress made in AI-based image generation~\cite{croitoru2023diffusion}. 
We provide an accessible toolbox that supports high-quality research into this developing field while increasing awareness for unrestricted AE research.
\section{Acknowledgments}
This work is supported by the Center for Digital Innovations of Lower Saxony (ZDIN) and
has received funding from the Lower Saxony Ministry of Science and Culture under grant numbers ZN3492 ("Zukunftslabor Gesellschaft \& Arbeit") and 51171389 ("CAIMed - Niedersächsisches Zentrum für KI \& Kausale Methoden in der Medizin") within the Lower Saxony “zukunft.niedersachsen” funding program of the Volkswagen Foundation.


\bibliography{main.bib}

\newpage
\appendix

\section{Appendix}

\end{document}